\documentclass[lettersize,journal]{IEEEtran}
\usepackage{amsmath,amsfonts}
\usepackage{algorithmic}
\usepackage{algorithm}
\usepackage{array}
\usepackage[caption=false,font=normalsize,labelfont=sf,textfont=sf]{subfig}
\usepackage{textcomp}
\usepackage{stfloats}
\usepackage{url}
\usepackage{verbatim}
\usepackage{graphicx}
\usepackage{cite}

\usepackage{bm}
\usepackage{multirow}
\usepackage{makecell}

\begin{document}

\title{Head-Tail Cooperative Learning Network for Unbiased Scene Graph Generation}


\author {Lei Wang, Zejian Yuan, Yao Lu, Badong Chen\IEEEauthorrefmark{1} 

\thanks{Lei Wang, Zejian Yuan and Badong Chen are from the Institute of Artificial Intelligence and Robotics, Xi'an Jiaotong University, Xi'an, 710049, China. (e-mail: leiwangmail@stu.xjtu.edu.cn, \{yuan.ze.jian, chenbd\}@mail.xjtu.edu.cn)}

\thanks{Yao Lu is from the Beijing Institute of Remote Sensing, Beijing, 100011, China. (e-mail: yaolu@bjirs.org.cn)}

\thanks{Corresponding Author: Badong Chen.}

}

\markboth{}%
{Shell \MakeLowercase{\textit{et al.}}: A Sample Article Using IEEEtran.cls for IEEE Journals}


\maketitle

\begin{abstract}
Scene Graph Generation (SGG) as a critical task in image understanding, facing the challenge of head-biased prediction caused by the long-tail distribution of predicates. However, current unbiased SGG methods can easily prioritize improving the prediction of tail predicates while ignoring the substantial sacrifice in the prediction of head predicates, leading to a shift from head bias to tail bias. To address this issue, we propose a model-agnostic Head-Tail Collaborative Learning (HTCL) network that includes head-prefer and tail-prefer feature representation branches that collaborate to achieve accurate recognition of both head and tail predicates. We also propose a self-supervised learning approach to enhance the prediction ability of the tail-prefer feature representation branch by constraining tail-prefer predicate features. Specifically, self-supervised learning converges head predicate features to their class centers while dispersing tail predicate features as much as possible through contrast learning and head center loss. We demonstrate the effectiveness of our HTCL by applying it to various SGG models on VG150, Open Images V6 and GQA200 datasets. The results show that our method achieves higher mean Recall with a minimal sacrifice in Recall and achieves a new state-of-the-art overall performance. Our code is available at https://github.com/wanglei0618/HTCL.
\end{abstract}

\begin{IEEEkeywords}
Image understanding, scene graph generation, head bias, tail bias, cooperative learning, self-supervised learning.
\end{IEEEkeywords}

\section{Introduction}
\begin{figure}[t]
\begin{center}
\includegraphics[width=0.45\textwidth]{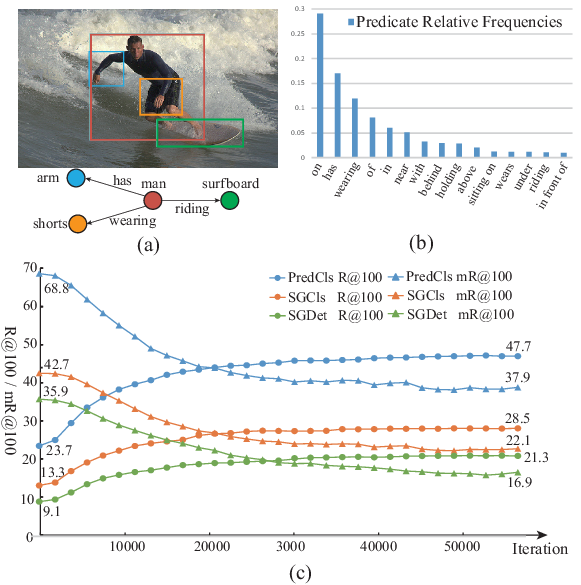}
\end{center}
   \caption{(a) Scene graph. (b) The long-tailed distribution of predicates in the Visual Genome (VG) dataset. (c) The SGG model experiences a shift from head bias to tail bias. With the fine-tuning of predicate classifier, the mRecall of head-biased SGG model improves substantially, while Recall decreases dramatically.}
\label{fig_intro}
\end{figure}

\IEEEPARstart{S} cene Graph Generation (SGG) is a critical image understanding task that involves detecting visual objects and their relationships in an image, as illustrated in Figure \ref{fig_intro}(a). The output is a structured $\langle$subject-predicate-object$\rangle$ triplet that represents the semantic information present in the image. This representation can facilitate various downstream tasks, such as image captioning \cite{caption2,captioning,huang2020image,zhang2018more,yu2018topic}, visual question answering \cite{VQA, vqa2, guo2021re, li2023weakly}, content-based image retrieval \cite{image_retrieval,image_retrieval2, song2018binary,bian2009biased,murala2012local,dubey2016multichannel}, human-object interaction detection \cite{He_2021_ICCV}, and image generation \cite{image_generation, image_generation2}.

However, current SGG methods face a significant challenge in terms of head-biased predictions, which arise from the long-tailed distribution of predicates in the dataset. As shown in Figure \ref{fig_intro}(b), commonly used datasets such as Visual Genome (VG) \cite{VG} are heavily skewed toward a few head predicate classes, while annotations for most tail predicate classes are scarce. As a result, coarse-grained head classes tend to dominate the predictions, while fine-grained tail classes that convey more specific semantic information are difficult to predict. 

Several techniques have been developed to mitigate head bias in SGG, including causal analysis \cite{tang2020_TDE}, loss re-weighting \cite{yan2020_pcpl,yu2020cogtree}, data re-sampling \cite{DT2, BGNN} and feature generation \cite{apfg}. Furthermore, the SGG evaluation metric has evolved from total sample-based Recall to mean Recall (mRecall) across all predicate classes.
However, improvements in the tail performance of SGG inevitably compromise head performance. If the substantial decline in head performance is overlooked while pursuing enhancements in the tail, SGG may experience a shift from head bias to tail bias.
To further demonstrate this issue, we propose a simple trick that focuses only on tail classes prediction (mRecall). Specifically, we fine-tune the last layer of a head-biased SGG model, i.e., the linear predicate classifier, on a class-balanced predicate sub-dataset based on re-sampling. As depicted in Figure \ref{fig_intro}(c),  this fine-tuning improves the model's mR@100 by 24.0, 15.2, and 12.2 for the three tasks, respectively, meeting or surpassing many carefully designed unbiased SGG methods \cite{tang2020_TDE,yan2020_pcpl,yu2020cogtree,BGNN, apfg,RTPB,DT2,SHA_GCL,BA_SGG}. However, the model's R@100 decreases by 30.9, 20.6, and 19.0, respectively. The underlying cause of these high mRecall but poor Recall methods is their tendency to allocate more head samples to tail classes, leading to a shift from head bias to tail bias. Hence, unbiased SGG methods should strive to minimize bias in both head and tail classes and consider both mRecall and Recall metrics when evaluating performance.

Therefore, this paper proposes a model-agnostic Head-Tail Cooperative Learning (HTCL) network for unbiased SGG, which aims to achieve higher mRecall while minimizing the sacrifice of Recall.  HTCL consists of two branches: a Head-Prefer Branch (HP-Branch) and a Tail-Prefer Feature Representation Branch (TPFR-Branch). The former is designed to recognize head class predicates and given an initial prediction, while the latter is responsible to represent the tail-prefer predicate feature and classify tail class predicates. The two branches of HTCL generate head-prefer and tail-prefer predictions respectively, and collaboratively implement the unbiased predicate prediction by a set of learnable class weight parameters.

To enhance the classification of rare tail predicates in the TPFR-Branch, we also propose a self-supervised learning approach to constrain tail-prefer predicate features. Specifically, we employ self-supervised contrast learning to maximize the dissimilarity between all predicate features, and then apply a head class center loss to encourage the head predicate features to converge to their respective class centers. This results in the aggregation of head samples and dispersion of tail samples, thereby reducing the difficulty of classifying tail predicates.

Our proposed model-agnostic HTCL network has been integrated into various classic SGG models on VG150, Open Images V6 and GQA200 datasets, including Motifs \cite{motifs}, Vctree \cite{tang2019_VCTREE}, Transformer, and PENet \cite{PENET}. Among them, the head-biased model termed HPC-ft, where we solely utilize the HP-Branch and fine-tune the classifier, achieved the best mRecall. Furthermore, our unbiased HTCL approach realized excellent mRecall while maintaining a high Recall, reaching a new state-of-the-art overall performance. This indicates that the unbiased scene graph generated by our HTCL method is not biased towards either head classes or tail classes. The main contributions of this paper are summarized below:
\begin{itemize}
\item This work show that unbiased SGG methods focus on boosting mRecall while ignoring the sacrifice of Recall, which may lead to a shift of SGG model from head bias to tail bias.
\item A head-tail cooperative learning network is proposed for unbiased SGG that cooperates with the predictions of head-prefer and tail-prefer, achieving optimal overall performance.
\item A self-supervised learning approach is introduced to the tail-prefer feature representation branch that improves the classification of tail predicates by constraining predicate features using self-supervised contrast learning and head center loss.
\end{itemize}

\section{Related Work}

\subsection{Unbiased Scene Graph Generation}

Early work \cite{KERN,tang2019_VCTREE,IMP,motifs,chen2022multi} on scene graph generation focused on developing model architectures and contextual feature fusion strategies. However, these methods suffered from head-biased predictions due to the inherent imbalance of predicates. In recent years, several unbiased approaches \cite{RTPB, DT2,SHA_GCL,BA_SGG,BGNN,ppdl,ebm,tang2020_TDE,yan2020_pcpl,yu2020cogtree,li2023uncertainty,lyu2023generalized,he2022towards,lyu2023adaptive,gao2023informative} have been proposed to address this problem. For example, TDE \cite{tang2020_TDE} uses causal analysis to eliminate prediction bias, Cogtree \cite{yu2020cogtree} employs cognitive tree loss for unbiased prediction, while BGNN \cite{BGNN} balances head and tail predicates through bi-level re-sampling. Moreover, the evaluation metric has evolved from total sample-based Recall to mRecall across all predicate classes. 
However, current unbiased SGG methods can easily prioritize achieving high mRecall over Recall, resulting in a significant decrease in Recall. This implies that the scene graphs generated by them are not truly unbiased, but rather shift from head bias to tail bias. In this work, we propose a head-tail cooperative learning network, with the aim of achieving higher mRecall while minimizing the sacrifice of Recall. As a result, we enable head and tail unbiased scene graph generation.

\subsection{Imbalanced Learning}
Imbalanced learning aims to tackle the long-tail recognition problem \cite{kang2019decoupling,zhou2020bbn,guo2021loss,xia2023generative,liu2022convolutional} caused by imbalanced data distribution, which can be mainly divided into data re-balancing and loss re-weighting. Data re-balancing involves adjusting the distribution of training data, typically by either oversampling or undersampling. Oversampling can be implemented using duplicate samples \cite{zhou2020bbn} or synthetic data \cite{xian2019f_vaegan}. In contrast, loss re-weighting assigns higher weights to the less frequent classes during training, allowing the classifier to focus more on them. These weights can be determined using prior knowledge, such as the class-balanced weight \cite{cb_loss}, which uses the inverse effective number of samples.
In this work, we implement re-weighting of the final predicate prediction to offset the imbalance in the training data. Additionally, we construct a balanced predicate features set using data re-sampling and fine-tune the tail-prefer classifier in the cooperative learning network. With our proposed method, the recognition of tail predicates in scene graphs can be improved.

\section{Biases in SGG: Motivation}

\begin{figure}[t]
\begin{center}
\includegraphics[width=0.45\textwidth]{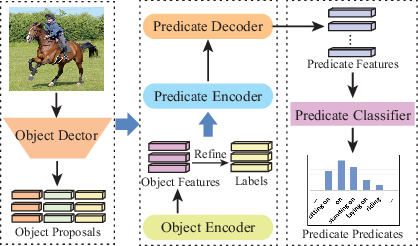}
\end{center}
   \caption{Illustration of the pipeline of classical SGG Models.}
\label{fig_sgg_pipeline}
\end{figure}

The most current unbiased SGG research is dedicated to addressing the bias towards head classes caused by imbalances in training data, with the aim of improving the predication of tail samples. However, in an unbiased SGG, enhancing the performance for tail classes inevitably compromises the head classes. If an unbiased SGG only pursues improvement in tail performance at the expense of head performance, the SGG bias would shift from being head-biased to tail-biased, resulting in the SGG model that is not genuinely unbiased.

As illustrated in Figure \ref{fig_sgg_pipeline}, the classical SGG pipeline can be divided into three components: extraction of object proposals from an image, representation of the predicate features, and classification of the predicates. 
In the SGG training set, the distribution of relationships across different classes is typically imbalanced. As illustrated in Figure \ref{fig_intro}(b), there is a pronounced skew toward a few head predicate classes, while the annotations for the tail predicate classes are sparse. Consequently, coarse-grained head classes are more likely to predominate in predictions, making it challenging to predict fine-grained tail classes that impart more specific semantics. This phenomenon is called head bias in SGG.

However, in unbiased SGG, focusing solely on the performance enhancement of tail classes while neglecting the sharp decline of head classes could potentially lead to a shift from head bias to tail bias. To validate this, we first trained a head-biased model on the imbalanced VG150 dataset using the classic SGG model. Subsequently, we constructed a class-balanced training set based on predicate features by re-sampling and fine-tuning the predicate classifier, resulting in a tail-biased model.

\begin{figure}[t]
\begin{center}
\includegraphics[width=0.4\textwidth]{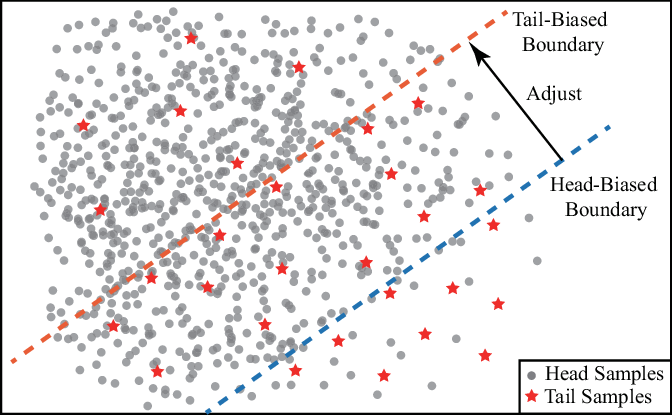}
\end{center}
   \caption{Illustration of the head-biased and tail-biased boundaries.}
\label{fig_boundaries}
\end{figure}

In the results presented in Table \ref{table_bias}, the head-biased model achieves the highest Recall representing head performance on three SGG tasks. On the contrary, the tail-biased model excels in two tasks, achieving the optimal mRecall indicative of tail performance, outperforming many unbiased SGG methods. We believe that fine-tuning on the predicate classifier corresponds to adjusting the decision boundaries for predicate classification. The tail-biased model assigns more samples to tail classes to enhance tail performance, as depicted in Figure \ref{fig_boundaries}. 
This leads to a significant number of head samples being misclassified as tail samples, causing the SGG model to shift from head bias to tail bias. Furthermore, we observed that DT2-ACBS \cite{DT2} and SHA+GCL \cite{SHA_GCL} also showed high mRecall but low Recall, indicating a risk of tail bias.

Thus, we argue that solely pursuing improvements in the performance of tail classes cannot lead to a genuinely unbiased SGG. An unbiased SGG should consider both the head and tail classes' performances. Therefore, we propose a head-tail cooperative learning network for SGG, which improves tail performance while minimally compromising head performance.

\begin{table}\small
\caption{Recall@100 and mRecall@100 of head-biased and tail-biased models on VG150.}
\label{table_bias}
\setlength{\tabcolsep}{2pt}
\centering
\begin{tabular}{l|cc|cc|cc}\hline
&\multicolumn{2}{c|}{PredCls}&
\multicolumn{2}{c|}{SGCls}&
\multicolumn{2}{c}{SGDet}\\
Models &Recall &mRecall  
&Recall &mRecall 
&Recall &mRecall \\
\hline
&   & &   &     & &   \\[-9pt]

Head-Biased &\textbf{68.8} & 23.7
&\textbf{42.7} & 13.3
& \textbf{35.9}& 9.1
\\

DT2-ACBS \cite{DT2} &25.6 & 39.7
&17.6 & 27.5
& 16.3& \textbf{24.4}
\\

SHA+GCL\cite{SHA_GCL} & 44.5 & 42.7
&22.2 & 21.3
& 17.9& 20.1
\\

Tail-Biased &37.9 & \textbf{47.7}
&22.1 & \textbf{28.5}
& 16.9& 21.3
\\

\hline
\end{tabular}
\end{table}

\section{Head-Tail Cooperative Learning Network}

\begin{figure*}
\begin{center}
\includegraphics[width=0.95\textwidth]{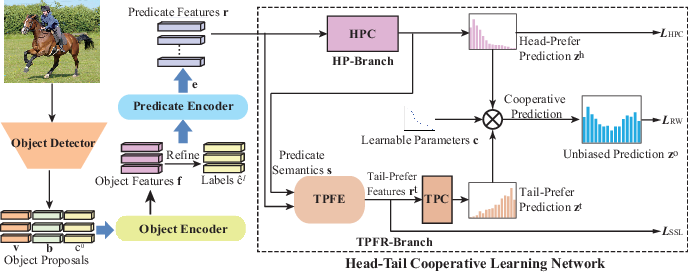} 
\end{center}
\caption{Overall pipeline of our head-tail cooperative learning network, where the Head-Prefer Branch (HP-Branch) and the Tail-Prefer Feature Representation Branch (TPFR-Branch) conduct unbiased predicate prediction by cooperating together. HPC is the head-prefer classifier, TPC is the tail-prefer classifier, and TPFE is the tail-prefer feature encoder.}
\label{fig_overall_model}
\end{figure*}

Figure \ref{fig_overall_model} summarizes the architecture of our proposed method. The classical SGG pipeline is used first to extract object proposals and predicate feature representation. Then, the predicate features are inputted into our proposed Head-Tail Cooperative Learning (HTCL) network, where the head-prefer branch and the tail-prefer feature representation branch conduct unbiased predicate prediction by cooperating together.

\subsection{Problem Formulation} SGG aims to generate a scene graph $\mathcal {G}=\{\mathcal{O},\mathcal{R}\}$ from an input image $\mathcal{I}$. The scene graph consists of a set of $n$ objects $\mathcal{O}=\{o_i\}^{n}_{i=1}$ and a set of $m$ relationships $\mathcal{R} = \{r_k\}^{m}_{k=1}$ between pairs of objects. Each object $o_i$ is described by a bounding box $\textbf{b}_i$ and a class label $c^l_i$.The generation of a scene graph $\mathcal{G}$ can be formulated as the joint probability distribution:
\begin{equation} \label{Eq_1}
Pr(\mathcal{G}|\mathcal{I})=Pr(\mathcal{B}|\mathcal{I})Pr(\mathcal{C}|\mathcal{I},\mathcal{B})Pr(\mathcal{R}|\mathcal{I},\mathcal{B},\mathcal{C}),
\end{equation}
where $Pr(\mathcal{B}|\mathcal{I})$ represents object proposals generation from input image, $Pr(\mathcal{C}|\mathcal{I},\mathcal{B})$ and $Pr(\mathcal{R}|\mathcal{I},\mathcal{B},\mathcal{C})$ denote object classification and predicate prediction, respectively.


\subsection{Predicate Feature Representation}

As illustrated in Figure \ref{fig_sgg_pipeline}, SGG methods typically employ a pre-trained object detector to extract a set of object proposals from the image $\mathcal{I}$, which provides a visual feature $\textbf{v}$, bounding box coordinates $\textbf{b}$, and an initial object label prediction $c^{0}$ for each object proposal.

With the object proposals, the object features $\textbf{f}$ can be encoded by \textbf{Object Encoder (OE)}:
\begin{equation} 
\textbf{f} = \textrm{OE}([\textbf{v}, pos(\textbf{b}), emb(c^{0})]),
\end{equation}
where $[\cdot,\cdot]$ denotes the concatenation operation, $pos$ is a fully-connected layer for object position encoding, $emb$ is a pre-trained Glove language model to acquire the word embedding. 

To enhance the predicate feature, the \textbf{Predicate Encoder (PE)} is used to obtain the context-aware object feature $\textbf e$ as follows:
\begin{equation}
\textbf e =  \textrm{PE}([\textbf{v}, \textbf{f}, emb(\hat{c}^l)]),
\end{equation}
where $\hat{c}^l$ is the predicted object label calculated by $\textbf{f}$.

Then the predicate feature $\textbf r$ between subject feature $\textbf e_i$ and object features $\textbf e_j$ can be decoded by a \textbf{Predicate Decoder (PD)} :
\begin{equation}
\textbf r =  \textrm{PD}( [\textbf e_i, \textbf e_j, \textbf{u}_{ij}]),
\end{equation}
where $\textbf{u}_{ij}$ is the union visual feature between subject $o_i$ and object $o_j$.

\subsection{Head-Tail Cooperative Learning Network}
Our head-tail cooperative learning network consists of two branches: a Head-Prefer Branch (HP-Branch) and a Tail-Prefer Feature Representation Branch (TPFR-Branch). HP-Branch initially performs a head-prefer prediction of the predicate features. Subsequently, both the initial predicate features and the head-prefer prediction are inputted into TPFR-Branch for a tail-prefer representation of the predicate features and a tail-prefer prediction. Finally, an unbiased prediction of the predicate is achieved by combining these two predictions using learnable weight parameters.

\textbf{HP-Branch:}
A Head-Prefer Classifier (HPC) is used to classify the predicate features $\textbf r$ from the predicate decoder. The HPC generates head-prefer prediction $\textbf z^h$ as follows:
\begin{equation}
\textbf z^h= W^h_{cls}\textbf{r} = [z^h_1, z^h_2,...,z^h_C]^T.
\end{equation}
Here, $W^h_{cls}$ denotes the HPC parameters and $C$ is the number of classes.

\textbf{TPFR-Branch:} 
To obtain the semantic representation of head-prefer prediction to input to TPFR-Branch based on the $\langle$subject-predicate-object$\rangle$ structure, we use the following formula:
\begin{equation}
\textbf s= [emb(\hat{c}_i^l),emb(\bm \sigma(\textbf z^h)),emb(\hat{c}_j^l)],
\end{equation}
where $\bm \sigma$ is the softmax function.

Then, a Tail-Prefer Feature Encoder (TPFE) is proposed to enhance the discriminability of the tail predicate features. TPFE consists of stacked transformer encoders \cite{Transformer} that use the self-attention mechanism to encode the relevance between all predicates in an image, thereby generating tail-prefer predicate features $\textbf{r}^{t}$ as follows:
\begin{equation}
\textbf{r}^{t} =  \textrm{TPFE}([[\textbf s_{1},\textbf r_{1}],...,[\textbf s_{m},\textbf r_{m}]]).
\end{equation}
Based on the $\textbf{r}^{t}$, a Tail-Prefer Classifier (TPC) is introduced to identify predicate classes, where the tail-prefer prediction $\textbf z^t$ are defined as:
\begin{equation}
\textbf z^t= W^t_{cls}\textbf{r}^{t} = [z^t_1, z^t_2,...,z^t_C]^T,
\end{equation}
where $W^t_{cls}$ refers to the parameters of TPC.

\textbf{Head-Tail Cooperative Prediction:} 
To fully harness the respective strengths of HP-Branch and TPFR-Branch, enabling each to handle the classifications of head and tail classes they excel at, this study defines a set of learnable class weight parameters $\textbf c$, denoted as:
\begin{equation}
\textbf c= [c_1, c_2,...,c_C]^T,
\end{equation}
Subsequently, the unbiased predicate prediction ${\textbf z}^o$ based on the head-prefer prediction $\textbf z^h$ and tail-prefer prediction $\textbf z^t$ can be expressed as:
\begin{equation}
{\textbf z}^o = N(\textbf z^h) \text{Sigmoid}(\textbf c)^T + N(\textbf z^t)(1-\text{Sigmoid}(\textbf c))^T
\label{eq_cooperative}
\end{equation}
Here, $N(\cdot)$ represent normalization operation, $\text{Sigmoid}(\cdot)$ is sigmoid function. Finally, the predicate label is calculated as $\hat{\textbf p}= \bm \sigma(\textbf z^o)$.

\begin{figure}[t]
\begin{center}
\includegraphics[width=0.35\textwidth]{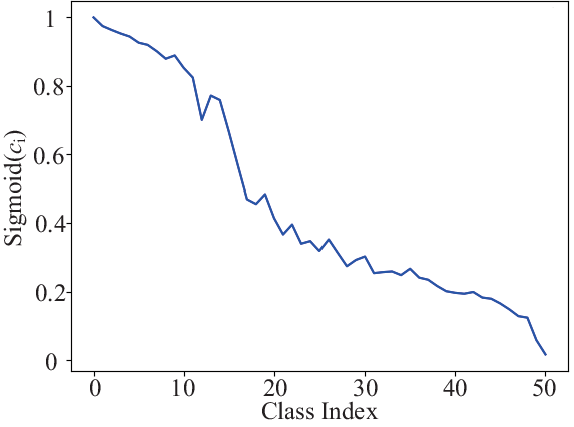}
\end{center}
   \caption{Learnable class weight parameters $\textbf c$, $\text{Sigmoid}(c_i)$ sorted in descending order according to the frequency of class samples.}
\label{fig_weight}
\end{figure}

To validate that the class weight parameter $\textbf c$ can harness the strengths of both HP-Branch and TPFR-Branch, Figure \ref{fig_weight} presents $c_i$ sorted in descending order according to the frequency of class samples after training. As observed, for high-frequency head classes, the class weight $\text{Sigmoid}(c_i)$ approaches 1, indicating that the prediction of the head classes in HTCL predominantly relies on the prediction of HP-Branch. However, as the sample frequency of the classes decreases, the class weight gradually declines, indicating an increasing dependence of HTCL on TPFR-Branch for the prediction of tail classes. Consequently, HTCL achieves collaborative prediction based on HP-Branch and TPFR-Branch through learnable weight parameters $\textbf c$.

\subsection{Tail-Prefer Feature Self-Supervised Learning} 

To enhance the discriminative capability of TPFR-Branch for tail samples, we utilize TPFE to re-represent the predicate features based on initial predicate features $\textbf{r}$ and the semantic information $\textbf{s}$ predicted by HP-Branch, with the aim of obtaining tail-prefer predicate features $\textbf{r}^{t}$.

\begin{figure}[t]
\begin{center}
\includegraphics[width=0.45\textwidth]{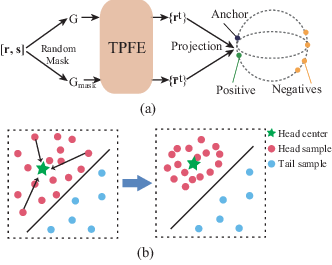}
\end{center}
   \caption{Illustration of tail-prefer feature self-supervised learning. (a) Contrast learning flow based on Tail-Prefer Feature Encoder (TPFE). (b) Description of head center loss.}
\label{fig_ssl_loss}
\end{figure}

To differentiate the few tail samples from the abundant head samples, we first introduce a self-supervised contrastive learning loss $\mathcal L_{Con}$ that is not affected by class labels to increase the distance between all samples, as illustrated in Figure \ref{fig_ssl_loss}(a). Subsequently, we employ head center loss $\mathcal L_{HC}$ to encourage all head samples to converge toward their class centers, as shown in Figure \ref{fig_ssl_loss}(b), thus improving the discriminability of head samples in the feature space.

For each image, the set of all its predicate representations $G = \{\textbf g_{k} |\textbf g_{k} =  \textrm{MLP}[\textbf r_k,\textbf s_k]\}_{k=1}^m$ can be obtained by an MLP. To implement contrast learning, an augmented set ${G}_{\textrm{mask}} = \{p_m(\bar{\textbf g}_{k}, \textbf g_{k})\}_{k=1}^m$ based on random perturbations is created by masking each $\textbf g_k$ with probability $p_m$ to the mean feature $\frac{1}{m}\sum_{k=1}^m \textbf g_{k}$. Furthermore, the tail-prefer predicate feature set and its augmented set are obtained by $\{\textbf{r}^{t}_k\}_{k=1}^m = \textrm{TPFE}(\textrm{G})$ and $\{\bar{\textbf{r}}^{t}_k\}_{k=1}^m =  \textrm{TPFE}(\textrm{G}_{\textrm{mask}})$ respectively. These features are then mapped to a hypersphere to calculate contrast loss, resulting in a set of samples $A = \{\textbf q_i\}_{i=1}^{2m}$. For each anchor sample $\textbf q_i$, the corresponding positive sample $\textbf q_{j(i)}$ describes the same predicate as $\textbf q_i$ in the image. The remaining $2m-2$ samples describing other predicates are negative samples. The self-supervised contrastive learning loss \cite{Sim_con, supcon} is defined as follows:
\begin{equation}
\mathcal L_{Con} = -\sum_{\textbf q_i \in A} \textrm{log}\frac{\textrm{exp}(\textbf q_i \cdot \textbf q_{j(i)}/\tau)}{\sum_{\textbf q_a \in A(i)}\textrm{exp}(\textbf q_i\cdot \textbf q_a/\tau)}.
\end{equation}
Here, the $\cdot$ symbol denotes the inner product, $\tau$ is a scalar temperature parameter and $A(i)\equiv A\setminus\{i\}$. As shown in Figure \ref{fig_ssl_loss}(a), $\textbf q_i \cdot \textbf q_{j(i)}$ in the numerator implies that the anchor and the positive sample are brought closer together, while $\textbf{q}_i\cdot \textbf{q}_a$ in the denominator implies that the anchor is kept away from all negative samples.

Head center loss \cite{center_loss} $\mathcal L_{HC}$ aims to reduce the volume occupied by head samples in feature space by clustering them toward their class centers, as illustrated in Figure \ref{fig_ssl_loss}(b). The loss function is defined as follows:
\begin{equation}
\mathcal L_{HC} = \sum_{i\in \mathcal{D}_h} ||\textbf{r}^t_i-C_{y_i}||_2,
\end{equation}
where $C_{y_i}$ is the mean feature of corresponding class $y_i$ and $\mathcal{D}_h$ is the head sample set in mini-batch. 

Thus, the self-supervised loss to tail-prefer predicate features is defined as follows:
\begin{equation}
\mathcal L_{SSL} = \mathcal L_{Con} + \lambda \mathcal L_{HC},
\end{equation}
where $\lambda$ is a hyperparameter used to regulate $\mathcal L_{HC}$.

\subsection{Training Loss} 
To offset the class imbalance in the dataset, we re-weight cross entropy loss of  the unbiased predicate prediction according to the sample size of different classes as follows:
\begin{equation}
\mathcal L_{RW} = -\sum_{i=1}^C w_i p_i \textrm{log}(\hat{p}_i),
\end{equation}
where $\hat{p}_i$ and $p_i$ are prediction and label, respectively. The weight of each class $i$ is denoted by $w_i$, which is used to adjust the impact of that class on the model based on its frequency. $w_i$ is calculated based on the effective number of samples in each class \cite{cb_loss} as follows $w_i = {(1-\beta)}/{(1-\beta^{n_i})}$, where $n_i$ is the number of samples in class $i$ and $\beta$ is a hyperparameter.

We also employ $\mathcal L_{HPC}$ to improve the classification of head classes while allowing the predicate feature representation model to learn robust feature representations from massive head class samples. The loss of HPC is defined as
\begin{equation}
\mathcal L_{HPC} =  \textrm{CE}( \textbf{p},\hat{ \textbf{p}}^h),
\end{equation}
where CE is the standard cross-entropy function, the estimated distribution of HPC is denoted by $\textbf{p}^h = \sigma(\textbf z^h)$.

In summary, the total training loss is 
\begin{equation}
\mathcal L_{\textrm{total}} = \mathcal L_{SSL}  +  \mathcal L_{RW} + \mathcal L_{HPC}  + \mathcal L_{obj},
\end{equation}
where $\mathcal L_{obj} = \textrm{CE}(\textbf c^l,\hat{\textbf c}^l)$ is the object classification loss in the predicate feature representation.


\section{Experiments}
\subsection{Dataset}
\textbf{Visual Genome (VG):} Following previous works \cite{RTPB, KERN, DT2,SHA_GCL,BA_SGG,BGNN,ppdl,ebm,tang2020_TDE,tang2019_VCTREE,IMP,yan2020_pcpl,yu2020cogtree,motifs,MLMG-SGG,chen2022multi} , our proposed method and recent methods are evaluated on the widely used subset of the VG dataset (that is, VG150) \cite{VG}, which consists of the most frequent 150 object classes and 50 predicate classes. Then we divide it into 70\% training set, 30\% testing set, and 5k images selected from the training set for validation.

\textbf{Open Images (OI):} The Open Images dataset \cite{open_image} is a large-scale dataset proposed by Google, which provides superior annotation quality for SGG. In this work, we conduct experiments on Open Images V6 (OIv6), follow the data processing and evaluation protocols in \cite{BGNN,gps_net,RelDN,PENET}. OIv6 has 126,368 images used for training, 1813 and 5322 images for validation and test, respectively, with 301 object classes and 31 predicate classes.

\textbf{Generalized Question Answering (GQA):} GQA \cite{VQA} is another vision and language benchmark with more than 3.8M relation annotations. We conduct experiments on GQA200 following data processing in \cite{SHA_GCL}, which consists of Top-200 object classes and Top-100 predicate classes. Similarly to VG150, GQA200 uses 70\% of the images for training, the remaining 30\% for testing and sample a 5K validation set from the training set.

\subsection {Task \& Evaluation Metrics} 
The SGG involves three subtasks \cite{IMP}:
1) Predicate Classification (\textbf{PredCls}) predicts the pairwise predicates with ground-truth object labels and bounding boxes.
2) Scene Graph Classification (\textbf{SGCls}) predicts both the object labels and their pairwise predicates with the ground-truth bounding boxes.
3) Scene Graph Detection (\textbf{SGDet}) detects all objects in an image and predicts their bounding boxes, labels, and predicates.

Although the initial evaluation metric used in SGG was total sample-based Recall@K (R@K), it was found to be dominated by head classes due to the imbalanced distribution of predicates in the dataset. To address this issue, \cite{tang2020_TDE} proposed using the mean Recall@K (mR@K) across all predicate classes to evaluate SGG. However, focusing solely on mR@K and neglecting R@K can result in a shift from head bias to tail bias in SGG. Therefore, some work \cite{DHL,DBiased,PENET,DCNET} utilizes the mean M@K of R@K and mR@K as an overall metric to jointly evaluate SGG. However, M@K tends to overvalue models with high R@K but low mR@K, often seen in models with head biases (initial models). Consequently, this paper adopts the harmonic mean F@K, which gives greater weight to smaller values.

\subsection{Implementation Details} 
We utilized the widely adopted pre-trained Faster RCNN with ResNeXt-101-FPN \cite{tang2020_TDE} as the object detector. For the HTCL, we select the top $h=10$ classes as the head class set. The hyperparameter $\beta$ that governs the re-weight loss $\mathcal L_{RW}$ is set to 0.9999. The initial class weight parameter $c_i$ is the logarithm of the number of predicates for the corresponding class.
For TPFE, we employ four transformer encoder layers, set the probability threshold $p_m$ for the augmented set ${G}_{\textrm{mask}}$ to 0.1, the scalar temperature parameter $\tau$ for contrast learning loss $\mathcal L_{Con}$ to 0.1, and the hyperparameter to regulate head center loss $\mathcal L_{HC}$ to $1 \times 10^{-4}$.
For classifier fine-tuning, the number of resamples for each predicate class is set to 5,000.
The models are implemented on a NVIDIA 3090 GPU with batch size 8 and learning rate $1 \times 10^{-3}$.

\subsection{Comparison with State of the Arts}

\begin{table*}\small
\caption{Comparison between our method and existing methods on the VG150 dataset. \dag  denotes our reproduced model. HPC-ft is the predicate classifier fine-tuning with only the head prefer branch using a predicate class-balanced training set.}
\label{table_results}
\setlength{\tabcolsep}{2pt}
\centering
\begin{tabular}{l|ccc|ccc|ccc}\hline
&\multicolumn{3}{c|}{PredCls}&
\multicolumn{3}{c|}{SGCls}&
\multicolumn{3}{c}{SGDet}\\
Models &R@50/100 &mR@50/100 &F@50/100 
&R@50/100 &mR@50/100 &F@50/100 
&R@50/100 &mR@50/100&F@50/100 \\
\hline
&   & &   &     & &  & &   & \\[-9pt]

KERN \cite{KERN} 
&\textbf{65.8/67.6} &17.7/19.2& 27.9/29.9 
&36.7/37.4 &9.4/10.0& 15.0/15.8 
&27.1/29.8&6.4/7.3&  10.4/11.7 
\\

PCPL \cite{yan2020_pcpl}
&50.8/52.6 &35.2/37.8& 41.6/44.0 
&27.6/28.4&18.6/19.6 & 22.2/23.2 
&14.6/18.6&9.5/11.7& 11.5/14.4 
\\

BGNN \cite{BGNN} 
& 59.2/61.3&30.4/32.9& 40.2/42.8 
&\textbf{37.4/38.5} &14.3/16.5& 20.7/23.1 
&\textbf{31.0/35.8}&10.7/12.6& 15.9/18.6 
\\

DT2-ACBS \cite{DT2}
&23.3/25.6 &35.9/39.7& 28.3/31.1 
&16.2/17.6 &\textbf{24.8/27.5}& 19.6/21.5 
&15.0/16.3&\textbf{22.0/24.4}& 17.8/19.5 
\\
SHA+GCL \cite{SHA_GCL} 
&42.7/44.5 &\textbf{41.0/42.7}& 41.8/43.6 
&21.4/22.2 &20.6/21.3& 21.0/21.7 
&14.8/17.9&17.8/20.1& 16.2/18.9 
\\

DTrans+RTPB\cite{RTPB}
&45.6/47.5 &36.2/38.1 & 40.4/42.3 
& 24.5/25.5&21.8/22.8& 23.1/24.1 
&19.7/23.4&16.5/19.0& 18.0/21.0 
\\

DCNET \cite{DCNET}
&57.3/59.1&33.4/35.6&\textbf{42.2/44.4}
&36.0/36.8&21.2/22.2&\textbf{26.7/27.7}
&28.6/32.9&14.3/17.3&\textbf{19.1/22.7}
\\

\hline
&   &  & & & &   & & \\[-9pt]
Motifs\dag \cite{motifs}  
& \textbf{65.7/67.9}  & 17.4/19.3  & 27.5/30.0 
& \textbf{41.3/42.5}  & 10.9/12.0  & 17.3/18.8 
& \textbf{32.0/36.3}  & 7.3/8.6  & 11.9/14.0  
\\
\quad+TDE \cite{tang2020_TDE} 
&46.2/51.4 &25.5/29.1& 32.9/37.2 
&27.7/29.9 &13.1/14.9& 17.8/19.9 
&16.9/20.3&8.2/9.8& 11.0/13.2 
\\

\quad+DeC \cite{DeC}
 & 59.2/60.6 & 18.3/20.3 & 28.0/30.4 
 & 34.6/35.9 & 11.8/12.3 & 17.6/18.3 
 & 27.7/30.8 & 9.0/10.4 & 13.6/15.5 
\\

\quad+TDE+DeC \cite{DeC}
 & 49.5/51.3 & 25.1/28.9 & 33.3/37.0 
 & 25.3/28.7 & 14.2/16.1 & 18.2/20.6 
 & 21.4/25.2 & 12.0/13.6 & 15.4/17.7
\\

\quad+BA-SGG\cite{BA_SGG}
&50.7/52.5 &29.7/31.7& 37.5/39.5 
& 30.1/31.0&16.5/17.5& 21.3/22.4 
&23.0/26.9&13.5/15.6& 17.0/19.7 
\\

\quad+CogTree\cite{yu2020cogtree}
&35.6/36.8 &26.4/29.0 &  30.3/32.4 
&21.6/22.2  &14.9/16.1& 17.6/18.7 
&20.0/22.1&10.4/11.8& 13.7/15.4 
\\

\quad+GCL \cite{SHA_GCL}
&42.7/44.4 &36.1/38.2&  39.1/41.1 
&26.1/27.1 &20.8/21.8&  23.2/24.2 
&18.4/22.0&16.8/19.3& 17.6/20.6 
\\

\quad+PPDL \cite{ppdl}
&47.2/47.6  &32.2/33.3 &38.3/39.2 
&28.4/29.3  &17.5/18.2&  21.7/22.5 
&21.2/23.9&11.4/13.5& 14.8/17.3 
\\

\quad+RTPB \cite{RTPB} 
&40.4/42.5 &35.3/37.7&37.7/40.0 
&26.0/26.9 &19.4/20.6& 22.2/23.3 
&19.0/22.5&13.1/15.5& 15.5/18.4 
\\

\quad+PCL \cite{PCL}
&55.0/57.3&33.6/35.8&41.7/44.1 
&34.2/35.2&18.2/19.1&23.8/24.8 
&29.0/33.4&14.2/16.6&19.1/22.2 
\\

\quad+DBiased \cite{DBiased}
&58.8/60.7&34.7/36.6&43.6/45.7 
&36.5/37.4&20.3/21.2&26.1/27.1 
&29.4/33.9&14.9/17.5&19.8/23.1 
\\

\quad+DHL \cite{DHL}
&51.8/53.8&39.1/41.7&44.6/47.0 
&27.4/31.1&23.1/24.2&25.1/27.2 
&24.7/28.8&\textbf{17.8/20.7}&\textbf{20.7/24.1}
\\

\textbf{\quad+HTCL (HPC-ft)}
& 32.8/34.7  & \textbf{41.1/43.6}  & 36.5/38.6 
& 20.9/22.0  & \textbf{24.7/26.3}  & 22.6/23.9 
& 14.6/17.2  & 17.4/20.1  & 15.9/18.5 
\\

\textbf{\quad+HTCL (Full)}
&59.3/61.1  & 36.6/39.1  & \textbf{45.2/47.7}
&37.1/37.9  & 23.0/24.4  & \textbf{28.4/29.7}
& 28.8/32.7 & 14.7/17.6 & 19.5/22.9 
\\
\hline
&  &    &  &   & &  & & \\[-9pt]
VCTree\dag \cite{tang2019_VCTREE}
& \textbf{66.3/68.4}  & 18.0/19.7  & 28.3/30.5 
& \textbf{46.6/47.9} & 12.1/13.2  & 19.1/20.7 
& \textbf{31.4/35.6} & 7.2/8.5  & 11.7/13.7 
\\

\quad+TDE \cite{tang2020_TDE} 
&47.2/51.6 &25.4/28.7& 33.0/36.9
& 25.4/27.9&12.2/14.0& 16.5/18.6 
&19.4/23.2&9.3/11.1&12.6/15.0 
\\

\quad+BA-SGG\cite{BA_SGG}
&50.0/51.8 &30.6/32.6& 38.0/40.0 
& 34.0/35.0&20.1/21.2& 25.3/26.4 
&21.7/25.5&13.5/15.7& 16.6/19.4 
\\

\quad+CogTree\cite{yu2020cogtree}
&44.0/45.4 &27.6/29.7 & 33.9/35.9 
&30.9/31.7 &18.8/19.9& 23.4/24.5 
&18.2/20.4&10.4/12.1& 13.2/15.2 
\\

\quad+GCL \cite{SHA_GCL} 
&40.7/42.7 &37.1/39.1& 38.8/40.8 
&27.7/28.7 &22.5/23.5& 24.8/25.8 
&17.4/20.7&15.2/17.5& 16.2/19.0 
\\

\quad+PPDL \cite{ppdl} 
&47.6/48.0  &33.3/33.8 & 39.2/39.7 
&32.1/33.0  &21.8/22.4 & 26.0/26.7 
&20.1/22.9&11.3/13.3 & 14.5/16.8 
\\

\quad+RTPB \cite{RTPB} 
&41.2/43.3 &33.4/35.6& 36.9/39.1 
&28.7/30.0 &24.5/25.8& 26.4/27.7 
&18.1/21.3&12.8/15.1& 15.0/17.7 
\\

\quad+PCL \cite{PCL}
&53.4/56.2&32.9/35.7&40.7/43.7 
&38.4/39.5&25.2/26.3&30.4/31.6 
&27.6/31.9&14.8/17.4&19.3/22.5 
\\

\quad+DBiased \cite{DBiased}
&59.1/61.0&34.5/36.4&43.6/45.6 
&36.8/37.7&20.4/21.3&26.2/27.2 
&29.5/34.1&14.3/17.0&19.3/22.7 
\\

\quad+DHL \cite{DHL}
&52.3/54.2&40.0/42.2&\textbf{45.3/47.5}
&36.6/37.8&26.9/28.2&31.0/32.3 
&23.3/27.1&\textbf{17.4/20.0}&19.9/23.0 
\\

\textbf{\quad+HTCL (HPC-ft)}
& 34.1/36.0  & \textbf{42.2/44.5}  & 37.7/39.8  
& 23.1/24.4  & \textbf{28.8/30.4}  & 25.7/27.1 
& 14.1/16.7 & 17.0/\textbf{20.0} & 15.4/18.2 
\\

\textbf{\quad+HTCL (Full)}
&58.8/60.5  & 36.5/39.1  & 45.0/\textbf{47.5}
&42.2/43.3  & 24.8/26.5  & \textbf{31.3/32.9} 
& 26.7/30.3 & 13.4/15.7 & 17.8/20.6
\\
\hline
& & & &  & & & & \\[-9pt]
Transformer\dag
& \textbf{66.7}/\textbf{68.8} &21.4/23.7 &    32.4/35.2 
& \textbf{41.7}/\textbf{42.7} & 12.2/13.3&   18.9/20.3 
& \textbf{31.6}/\textbf{35.9}&7.6/9.1	 &  12.3/14.5 
\\

\quad+PCL \cite{PCL}
&57.3/59.2&36.3/39.2&44.4/47.2 
&36.0/37.0&20.7/21.8&26.3/27.4 
&29.9/34.2&15.2/18.3&20.2/\textbf{23.8}
\\

\quad+DHL \cite{DHL}
&49.0/51.1&40.4/42.6&44.3/46.5 
&28.1/29.1&24.2/25.3&26.0/27.1 
&23.2/27.3&\textbf{18.2}/21.0&\textbf{20.4}/23.7
\\

\textbf{\quad+HTCL (HPC-ft)}
& 35.9/37.9 &\textbf{44.9}/\textbf{47.7} &   39.9/42.2 
&20.9/22.1  &\textbf{27.0}/\textbf{28.5}  &  23.6/24.9 
& 14.2/16.9 & \textbf{18.2}/\textbf{21.3}	 &   15.9/18.8 
\\

\textbf{\quad+HTCL (Full)}
&59.1/60.9  & 39.7/42.7  & \textbf{47.5/50.2} 
&37.1/38.0  & 23.3/24.8  & \textbf{28.6/30.0}
& 28.0/31.9 & 14.8/17.4 & 19.4/22.5 
\\
\hline
& & & &  & & & & \\[-9pt]
PENet \cite{PENET} & \textbf{64.9}/\textbf{67.2} & 31.5/33.8 &  42.4/45.0 
&\textbf{39.4}/\textbf{40.7} & 17.8/18.9 &  24.5/25.8 
& \textbf{30.7}/\textbf{35.2} & 12.4/14.5 &  17.7/20.5 
\\
\quad+Reweight \cite{PENET} & 59.0/61.4 & 38.8/40.7 &  46.8/49.0 
&36.1/37.3  & 22.2/23.5 &  27.5/28.8 
& 26.5/30.9 & 16.7/18.8 &  20.5/23.4 
\\
\textbf{\quad+HTCL (Full)}
&61.4/63.3  & \textbf{41.5/44.1}  & \textbf{49.5/52.0} 
&36.1/37.0  & \textbf{25.9/27.3}  & \textbf{30.2/31.4} 
&28.1/32.3  & \textbf{17.3/20.1}  & \textbf{21.4/24.8}
\\
\hline
\end{tabular}
\end{table*}

\paragraph{VG150} To evaluate our proposed method on VG150, our approach combines four classical SGG models to represent predicate features, namely Motifs \cite{motifs}, VCTree \cite{tang2019_VCTREE}, Transformer \cite{Transformer} and PENet \cite{PENET}, as shown in Table \ref{table_results}. The results of the state-of-the-art methods that are being compared are divided into various debiased methods \cite{tang2020_TDE, DeC, BA_SGG, yu2020cogtree, SHA_GCL, ppdl, RTPB, PCL, DBiased,DHL} on classical models and specific SGG models \cite{KERN, yan2020_pcpl, BGNN, DT2, SHA_GCL, RelDN, DCNET}.

When comparing different SGG models using the R@K metric, classical models that focus on feature representation and fusion, such as Motifs, VCTree, Transformer, and PENet, achieve optimal results. For instance, VCTree achieves 66.3/68.4, 46.6/47.9 and 31.4/35.6 for R@50/100 on three tasks, respectively. However, due to the imbalanced distribution of predicates, R@K is dominated by head classes, resulting in head-biased SGG by these models.

Therefore, current methods prefer to use mR@K to evaluate SGG and ensure that the generated scene graphs are unbiased. However, pursuing mR@K improvement can lead to a sacrifice in R@K. For instance, DT2-ACBS and SHA+GCL have only achieved 15.0/16.3 and 14.8/17.9 for R@50/100, respectively, despite achieving 22.0/24.4 and 17.8/20.1 for mR@50/100 on the SGDet task. This excessive sacrifice of R@K can shift SGG from head bias to tail bias. To further illustrate this issue, we fine-tuned the predicate classifier of the head-biased models (Motifs, VCTree, and Transformer) on a balanced training set, which substantially improved the mR@K. The resulting model is called HPC-ft, which only has the HP-Branch and fine-tunes the head-prefer classifier.

Comparing the experimental results, HPC-ft outperforms the most carefully designed unbiased SGG methods on mR@K, such as HPC-ft based on Transformer achieving 44.9/47.7, 27.0/28.5, and 18.2/21.3 for mR@50/100 on the three tasks, respectively. However, this improvement comes at the expense of R@K, with HPC-ft reducing R@50/100 by 30.8/30.9, 20.8/20.6, and 17.4/19.0. Therefore, unbiased SGG models should consider both mR@K and R@K to ensure that the generated scene graphs are unbiased in both head and tail. In this paper, we prefer to use the harmonic mean value F@K of R@K and mR@K to evaluate unbiased SGG models.

Considering the F@K metric, our HTCL achieved a consistent improvement in all baselines, where PENet+HTCL achieved 49.5/52.0, 30.2/31.4, and 17.3/20.1 on PredCls, SGCls, and SGDet tasks with respect to F@50/100, respectively. These results exceed almost all other methods. Therefore, HTCL maintains a high R@50/100 while raising mR@50/100 to the state-of-the-art level, indicating that the scene graphs generated by HTCL are unbiased towards both head and tail.

\begin{table}\small
\caption{Comparison between our method and existing methods on OpenImages V6.}
\label{table_OIV6}
\setlength{\tabcolsep}{2pt}
\centering
\begin{tabular}{l|cccccc}
\hline
Models&mR@50 &R@50 &F@50 &wmAP$_{\text{rel}}$ &wmAP$_{\text{phr}}$ & score$_{\text{wtd}}$
\\
\hline
& & & & &   &    \\[-9pt]
RelDN \cite{RelDN}&34.0&	73.1&	46.4 & 32.2&	33.4	&40.8 \\
G-RCNN \cite{G-RCNN}	&34.0	&74.5	& 46.7 &33.2	&34.2	&41.8 \\
GPS-Net \cite{gps_net}&	35.3&	74.8& 48.0	& 32.9	&34.0	&41.7 \\
BGNN \cite{BGNN}&	41.7	&75.0	& 53.6 &33.8	&34.9	&42.5 \\
\hline
&	 & & & && \\[-9pt]

Motifs \cite{motifs}&	32.7&	71.6&	44.9 & 29.9&	31.6	&38.9 \\
\quad+HTCL&	42.7&76.4 &54.8 &37.1 &38.0 &45.1 \\
\hline
&	 & & & && \\[-9pt]

VCTree \cite{tang2019_VCTREE}&	33.9&	74.1&46.5	& 34.2	&33.1	&40.2 \\
\quad+HTCL&	36.4 & 75.2&49.1  &37.8 &38.7 &45.3 \\
\hline
&	 & & & &&  \\[-9pt]
Transformer&35.3	&76.4 &48.3 &\textbf{40.3} &\textbf{40.3} &\textbf{47.4} \\
\quad+HTCL & 42.1  & 76.1  & 54.2 & 39.1 & 39.6 & 46.5  \\
\hline
&	 & & & && \\[-9pt]
PENet \cite{PENET} & 39.3 &\textbf{76.7} &52.0 &37.1& 38.0 &45.2 \\
\quad+HTCL & \textbf{44.6} &76.4 & \textbf{56.3}&37.0& 38.0 &44.9 \\
\hline
\end{tabular}
\end{table}

\paragraph{Open Image V6} 
To validate the generalizability of HTCL, we performed experiments on Open Images V6 following the same evaluation protocols as in \cite{BGNN,gps_net,RelDN,PENET} on Table \ref{table_OIV6}. mR@50, R@50, weighted mean AP of relationships (wmAP$_{\text{rel}}$), and weighted mean AP of phrase (wmAP$_{\text{phr}}$) are used as evaluation metrics. Following standard Open Images evaluation metrics, the weight metric score$_{\text{wtd}}$ is calculated as: score$_{\text{wtd}}$ = 0.2 $\times$ R@50 + 0.4 $\times$ wmAP$_{\text{rel}}$ + 0.4 $\times$ wmAP$_{\text{phr}}$. Besides, we also report F@50 like Visual Genome as an overall metric for comprehensive comparison.

In line with VG150, when we applied HTCL to Motifs, VCTree, Transformer, and PENet, there is a consistent substantial increase in their tail performance at mR@50, with only a slight decrease in other metrics. For example, the values of PENet+HTCL in R@50, wmAP$_{\text{rel}}$, and score$_{\text{wtd}}$ decreased by 0.3, 0.1 and 0.3, respectively, while mR@50 saw an increase of 5.3. This demonstrates that HTCL significantly enhances tail performance while maintaining competitive performance at the head.

\begin{table*}\small
\caption{Comparison between our method and existing methods on GQA200.}
\label{table_gqa200}
\setlength{\tabcolsep}{2pt}
\centering
\begin{tabular}{l|ccc|ccc|ccc}\hline
&\multicolumn{3}{c|}{PredCls}&
\multicolumn{3}{c|}{SGCls}&
\multicolumn{3}{c}{SGDet}\\
Models &R@50/100 &mR@50/100 &F@50/100 
&R@50/100 &mR@50/100 &F@50/100 
&R@50/100 &mR@50/100&F@50/100 \\
\hline
&   & &   &     & &  & &   & \\[-9pt]
VTransE \cite{VTransE}& {55.7/57.9}& 14.0/15.0&  22.4/23.8 
& {33.4/34.2}& 8.1/8.7& 13.0/13.9 
&{27.2/30.7}& 5.8/6.6&  9.6/10.9 
\\

SHA+GCL \cite{SHA_GCL} & 42.7/44.5& \textbf{41.0/42.7} &{41.8/43.6}
&21.4/22.2 & \textbf{20.6/21.3} &{21.0/21.7}
&14.8/17.9 & \textbf{17.8/20.1}&{16.2/18.9}
\\

\hline
& & & &  & & & & \\[-9pt]

Motifs \cite{motifs} & \textbf{65.3/66.8}& 16.4/17.1&26.2/27.2 
& \textbf{34.2/34.9}& 8.2/8.6& 13.2/13.8 
& \textbf{28.9/33.1}& 6.4/7.7&10.5/12.5 
\\

\quad+GCL \cite{SHA_GCL}
& 44.5/46.2 & {36.7/38.1}&  40.2/41.8  
&  23.2/24.0 & {17.3/18.1}&  19.8/20.6 
& 18.5/21.8&{16.8/18.8} &  17.6/20.2
\\

\quad+HTCL 
& 54.5/56.3 & 34.1/35.4 & {42.0/43.5} 
& 28.8/29.6 & 17.2/17.7 & {21.6/22.2} 
& 23.5/27.1 & 16.0/18.0 & \textbf{19.0/21.6}
\\

\hline
& & & &  & & & & \\[-9pt]

VCTree \cite{tang2019_VCTREE}& {63.8/65.7}& 16.6/17.4 &26.3/27.5 
&{34.1/34.8} & 7.9/8.3 &12.8/13.4 
&{28.3/31.9} & 6.5/7.4 &10.6/12.0
\\

\quad+GCL \cite{SHA_GCL} 
& 44.8/46.6 &{35.4/36.7} &  39.5/41.1 
& 23.7/24.5 &{17.3/18.0} & 20.0/20.8 
&17.6/20.7 & 15.6/17.8& 16.5/19.1 
\\

\quad+HTCL 
& 56.6/58.3  & 32.6/33.9  & {41.4/42.9} 
& 28.9/29.7  & 15.7/16.3  & {20.3/21.0}
&22.6/25.8 & 14.0/15.8 & 17.3/19.6
\\

\hline
& & & &  & & & & \\[-9pt]
Transformer
&{65.2/66.7} & 19.1/20.2 & 29.5/31.1
&{33.9/34.6} & 9.3/9.7 & 14.6/15.2
&{27.4/31.5} & 6.7/7.9 & 10.7/12.6
\\

\quad+HTCL 
& 55.0/56.7  & {35.9/37.4}  & {43.5/45.1} 
& 28.1/28.9  & {18.3/19.0}  & \textbf{22.2}/22.9
& 22.5/26.0  & {14.7/17.1}  & {17.8/20.6} 
\\

\hline
& & & &  & & & & \\[-9pt]
PENET \cite{PENET}
&54.5/{56.2} & 27.3/28.1 & 36.4/37.5
&{27.3/28.1} & 12.2/12.6 & 16.9/17.4
&{22.1/26.1} & 11.1/12.8 & 14.8/17.2

\\

\quad+HTCL
&{55.2}/56.9 & {37.9/39.3} & \textbf{45.0/46.5}
&26.6/27.4 & {18.9/19.9} & 22.1/\textbf{23.1} 
&21.0/24.7 & {15.1/18.1} & {17.6/20.9} 
\\

\hline
\end{tabular}
\end{table*}

\paragraph{GQA200} 
We also applied HTCL to the more challenging GQA200, as shown in Table \ref{table_gqa200}. Compared to the GCL method \cite{SHA_GCL}, HTCL achieved a competitive mR@K while maintaining a higher R@K, resulting in optimal overall performance. For example, F@50 of PENet+HTCL for the three tasks are 45.0/46.5, 22.1/23.1, and 17.6/20.9, respectively. This confirms the generalizability of HTCL across different data distributions.

\subsection{Ablation Studies}

\begin{figure}
\begin{center}
\includegraphics[width=0.475\textwidth]{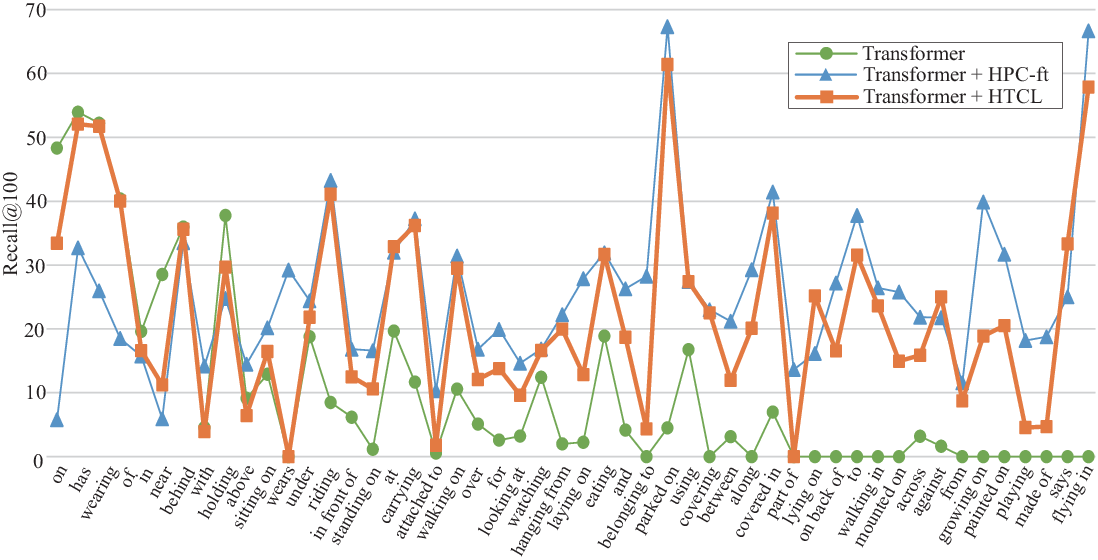} 
\end{center}
\caption{Comparison of the Recall@100 between head-biased model (Transformer), tail-biased model (Transformer+HPC-ft) and our HTCL unbiased model (Transformer+HTCL) for each predicate class of SGCls task on VG150. The frequencies of predicates decrease from left to right.}
\label{all_recall}
\end{figure}

\paragraph{Class-wise Performance Comparison}
To analyze the effectiveness of our proposed method in predicting each predicate class, we present a comparison of the Recall@100 for the head-biased model (Transformer), tail-biased model (Transformer+HPC-ft) and our HTCL unbiased
model (Transformer+HTCL) on all 50 predicate classes in Figure \ref{all_recall}, and the predicate classes are sorted by frequency.

The results show that the head-biased model performs well in head classes but struggles to accurately predict tail classes. This is due to the imbalanced dataset, where the top 10 predicate classes with the highest frequency account for 88.6\% of all samples. As a result, head-biased model only needs to accurately predict head predicates to achieve a high Recall. On the other hand, the tail-biased model focuses on predicting tail classes and ignores head classes, resulting in high mRecall but poor Recall. However, our proposed HTCL unbiased model is able to predict both head and tail classes effectively, achieving high Recall and mRecall simultaneously. This ensures that the generated scene graphs are not biased towards either the head or tail classes. 

\begin{table}\small
\caption{Ablation studies. Group A is based on different components of HTCL, Group B is based on the predictions of different classifiers in HTCL and Group C is based on the different components in the loss function. HP-Branch is head-prefer branch, TPFR-Branch is tail-prefer feature representation branch, TPFE is tail-prefer feature encoder, ft is predicate classifier fine-tuning, HPC is head-prefer classifier, TPC is tail-prefer classifier and $\mathcal{P}$ is denoted as predicate prediction based on different classifiers.}
\label{table_ablation_studies}

\setlength{\tabcolsep}{2pt}
\begin{center}
\begin{tabular}{cl|ccc}
\hline
&&\multicolumn{3}{c}{PredCls}\\
Group&Model &R@50/100&mR@50/100& F@50/100\\
\hline
& &  & &  \\[-9pt]
\multirow{4}{*}{A}&HP-Branch
& \textbf{66.7}/\textbf{68.8} &21.4/23.7 &    32.4/35.2 
\\
&HPC-ft
& 35.9/37.9 &\textbf{44.9}/\textbf{47.7} &   39.9/42.2 
\\
&TPFR-Branch 
&44.0/46.0 & 42.5/45.1 & 43.2/45.5 
\\
& w/o TPFE 
& 65.1/66.9 & 33.4/35.9 & 44.2/46.7 
\\
& w/o TPC-ft 
& 61.3/63.2 & 37.2/39.9 & 46.3/48.9 
\\
&${\textrm{HTCL}}$
& 59.1/60.9 & 39.7/42.7 & \textbf{47.5/50.2}

\\
\hline
&&  & &  \\[-9pt]
\multirow{4}{*}{B}&$\mathcal{P}_{\textrm{HPC}}$ 
& \textbf{67.1/69.0} & 23.0/25.1 & 34.2/36.8 
\\
&$\mathcal{P}_{\textrm{TPC}}$ 
& 9.8/11.5 & 39.4/42.1 & 15.7/18.1 
\\
&$\mathcal{P}_{\textrm{TPC w/o ft}}$ 
& 12.7/14.9 & 38.1/40.1 & 19.0/21.8 
\\
&$\mathcal{P}_{\textrm{HTCL}}$
& 59.1/60.9 & \textbf{39.7/42.7} & \textbf{47.5/50.2}
\\
\hline
&&  & &  \\[-9pt]
\multirow{6}{*}{C}&w/o $\mathcal{L}_{SSL}$ 
& 62.2/64.2 & 35.9/38.4 & 45.5/48.0 
\\
&w/o $\mathcal{L}_{Con}$ 
& 62.1/64.0 & 36.6/38.8 & 46.0/48.3  
\\
&w/o $\mathcal{L}_{HC}$ 
& 59.7/61.5 & 39.1/41.5 & 47.2/49.5 
\\
&w/o $\mathcal{L}_{HPC}$ 
& 57.9/59.9 & 39.5/41.7 & 47.0/49.2 
\\ 
&w/o $\mathcal{L}_{RW}$ 
& \textbf{62.5/63.6} & 21.2/23.3 & 31.7/34.1 
\\
&$\mathcal{L}_{\textrm{total}}$ 
& 59.1/60.9 & \textbf{39.7/42.7} & \textbf{47.5/50.2}
\\
\hline
\end{tabular}
\end{center}
\end{table}

\paragraph{Analysis of HTCL}
To assess the effectiveness of each component in HTCL, we performed ablation experiments in HTCL and presented the results in Table \ref{table_ablation_studies}. 
For the component analysis in Group A, the model with only HP-Branch and HPC-ft after fine-tuning in the balanced set achieved optimal results in R@50/100 and mR@50/100, respectively. However, they have head bias and tail bias respectively, which do not meet our demands. In contrast, TPFR-Branch, which only uses the tail-prefer feature representation branch, produces poor performance in head classes. Thus, both branches of HTCL have made significant contributions to unbiased prediction. The results without tail-prefer feature encoder (w/o TPFE) perform poorly in tail classes, indicating that the tail-prefer feature representation is crucial for HTCL to recognize tail samples. When comparing HTCL with w/o TPC-ft, we find that fine-tuning of the tail-prefer classifier helps improve the tail and overall performance.

For the full HTCL model, we report the prediction results of the head-prefer classifier $\mathcal{P}_{\textrm{HPC}}$ and the tail-prefer classifier $\mathcal{P}_{\textrm{TPC}}$ in Group B. The results clearly demonstrate that HPC and TPC are good at predicting head and tail classes, respectively, which aligns with our expectation for both branches of HTCL. Moreover, comparing $\mathcal{P}_{\textrm{TPC w/o ft}}$ with $\mathcal{P}_{\textrm{TPC}}$, we can see that fine-tuning TPC based on the predicate class-balanced training set improves its tail prediction performance.

\paragraph{Component Analysis of Training Loss} To assess the impact of different components of training loss, we performed ablation tests in Group C of Table \ref{table_ablation_studies}. Comparing w/o $\mathcal{L}_{SSL}$ with $\mathcal{L}_{\textrm{total}}$, the self-supervised loss $\mathcal{L}_{SSL}$ improves the model's ability to predict the tail classes. Furthermore, the results of w/o $\mathcal{L}_{Con}$ and w/o $\mathcal{L}_{HC}$ illustrate that contrast learning loss and head center loss improve the prediction of tail samples, respectively. The results of w/o $\mathcal{L}_{HPC}$ confirm that $\mathcal{L}_{HPC}$ can assist the predicate feature representation model to learn robust feature representations from massive head class samples, contributing to overall performance. Moreover, the results of w/o $\mathcal{L}_{RW}$ suggest that the re-weighted loss $\mathcal{L}_{RW}$ is crucial for predicting the tail classes.

\subsection{Hyperparameter Analysis}

\begin{table}\small

\caption{Parameter analysis of the re-weight hyperparameter $\beta$.}
\label{table_num_head}
\setlength{\tabcolsep}{2pt}
\begin{center}
\begin{tabular}{l|ccc}
\hline
&\multicolumn{3}{c}{PredCls}\\
$\beta$ &R@50/100&mR@50/100& F@50/100\\
\hline
 &  & &  \\[-9pt]
0.999& \textbf{62.2/63.7} & 36.7/39.6 & 46.1/48.8 \\
0.9995 & 61.5/63.1 & 39.5/42.4 & 48.1/50.7 \\
0.9999 & 61.4/63.3 & 41.5/44.1 & \textbf{49.5/52.0} \\
0.99995& 59.2/61.0 & 41.6/45.0 & 48.9/51.8  \\
0.99999& 52.8/55.0 & \textbf{42.4/45.5} & 47.0/49.8  \\
\hline

\end{tabular}
\end{center}

\end{table}

\paragraph{Re-weight Hyperparameter}
The re-weighting loss $\mathcal L_{RW}$ plays a crucial role in offsetting the imbalanced training set, and its hyperparameter $\beta$ significantly affects the model's performance. The analysis experiments for $\beta$ are presented in Table \ref{table_num_head}. As the value of $\beta$ increases, the weight of tail classes increases, resulting in a rapid decrease in the performance of head classes while improving the performance of tail classes. Based on the results, the optimal $\beta$ is chosen as 0.9999.

\begin{table}\small
\caption{Parameter analysis of the weight of head center loss $\lambda$}
\label{table_lambda}
\setlength{\tabcolsep}{2pt}
\begin{center}
\begin{tabular}{c|ccc}
\hline
&\multicolumn{3}{c}{PredCls}\\
$\lambda$ ($\times 10^{-4}$) &R@50/100&mR@50/100& F@50/100\\
\hline
 &  & &  \\[-9pt]
0.2
& \textbf{61.5/63.3} & 40.9/41.0 & 49.2/49.8 \\
0.5
& 60.8/62.7 & 40.7/43.7 & 48.7/51.5\\
1
& 61.4/\textbf{63.3} & 41.5/44.1 & \textbf{49.5/52.0}\\
2
& 60.7/62.4 & 41.4/44.0 & 49.2/51.6 \\
5
& 59.5/61.1 & \textbf{41.6/44.7} & 48.9/51.6\\
\hline
\end{tabular}
\end{center}

\end{table}

\paragraph{Weight of Head Center Loss}
The TPFR-Branch of HTCL utilizes the head center loss $\mathcal L_{HC}$ to encourage the clustering of head predicate features around their respective class centers, thus improving the discriminability of head-prefer predicate features. To evaluate the impact of the weight parameter $\lambda$ associated with $\mathcal L_{HC}$, we perform analysis experiments and present the results in Table \ref{table_lambda}. As $\lambda$ increases, the weight of $\mathcal L_{HC}$ increases, resulting in improved tail predicate classification performance. However, optimal overall performance is achieved when $\lambda$ is set to $1 \times 10^{-4}$. Consequently, we set $\lambda$ to this value in our experiments.

\begin{table}\small
\caption{Parameter analysis of the number of TPFE layers.}
\label{table_layers}
\setlength{\tabcolsep}{2pt}
\begin{center}
\begin{tabular}{c|ccc}
\hline
&\multicolumn{3}{c}{PredCls}\\
layers &R@50/100&mR@50/100& F@50/100\\
\hline
 &  & &  \\[-9pt]
2& \textbf{62.2/64.1} & 40.2/43.2 & 48.9/51.6 \\
3& 60.6/62.4 & 41.4/43.9 & 49.2/51.5 \\
4& 61.4/63.3 & 41.5/44.1 & \textbf{49.5/52.0}\\
5& 60.4/62.1 & 41.4/44.2 & 49.1/51.6 \\
6& 60.1/61.4 & \textbf{41.6/44.6} & 49.2/51.7 \\
\hline
\end{tabular}
\end{center}

\end{table}

\paragraph{Number of the TPFE layers}
In this study, we propose a Tail-Prefer Feature Encoder (TPFE) to re-represent predicate features and employ a self-supervised loss to reduce the classification difficulty of tail predicates. We conduct parameter analysis for the number of TPFE layers, and the results are summarized in Table \ref{table_layers}. Optimal overall performance is achieved when the TBE has 4 layers.

\section{Qualitative Study}

\begin{figure*}
\centering
\includegraphics[width=0.95\textwidth]{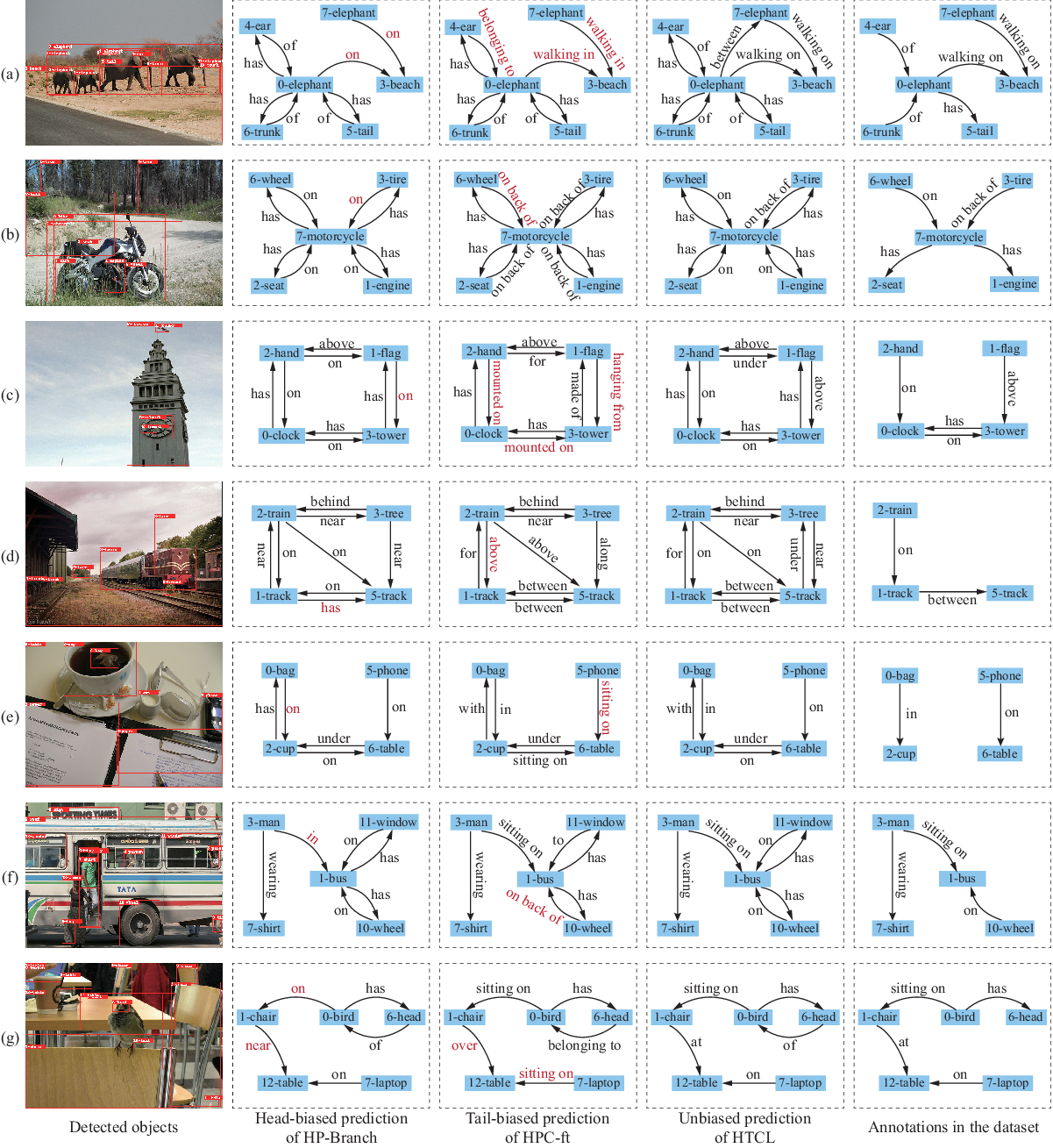} 
\caption{Visualized results of HP-Branch, HPC-ft and HTCL on the PredCls task. The red color indicates incorrect predicate predictions. Some trivial detected objects are excluded due to space limitation.}
\label{fig_visualize}
\end{figure*}

Figure \ref{fig_visualize} shows the visualized results of HP-Branch, HPC-ft, and HTCL on the PredCls task. The annotations of the VG dataset are sparse, but scene graph generation models can detect rich predicates. We can conclude the following:
\begin{itemize}
\item The head-biased prediction of HP-Branch struggles to identify tail predicates. For example, fine-grained predicates such as "walking on," "on back of," "above," and "sitting on" are all predicted as the head predicate "on." 

\item In contrast, tail-biased HPC-ft tends to predict more tail predicates than head predicates, such as $<$wheel-on back of-motorcycle$>$ and $<$laptop-sitting on-table$>$ in Figures \ref{fig_visualize} (b) and (g), respectively. 
 This confirms that the essence of HPC-ft is to divide more head predicates into tail classes, which would shift the SGG from head bias to tail bias.

\item Our proposed HTCL can predict both coarse-grained head predicates and fine-grained tail predicates, such as $<$wheel-on-motorcycle$>$ and $<$engine-on back of-motorcycle$>$ in Figure \ref{fig_visualize} (b), $<$laptop-on-table$>$ and $<$bird-sitting on-chair$>$ in Figure \ref{fig_visualize} (g).  This demonstrates that our proposed HTCL is capable of generating scene graphs that are neither head nor tail biased.
\end{itemize}

\section{Conclusion}
In this study, a head-tail collaborative learning network is proposed to achieve accurate recognition of head and tail predicates. The network consists of head-prefer and tail-prefer feature representation branches that collaborate with each other. Additionally, our proposed self-supervised learning approach improves the classification of tail samples by constraining the tail-prefer predicate features. The experimental results demonstrate that our method achieves a higher mean Recall with minimal sacrifice in Recall and delivers optimal overall performance.

\textbf{Limitation:} In line with most SGG methods, a limitation of our approach is its reliance on the performance of the pre-trained object detector, particularly in detecting object bounding boxes, which constrains the efficacy of HTCL in the SGDet subtask. Therefore, we will explore the application of HTCL in end-to-end SGG models in future work.

\section*{Acknowledgments}
This study was funded by the National Natural Science Foundation of China with grant numbers (U21A20485, 62311540022, 61976170).

\bibliographystyle{IEEEtran}
\bibliography{HTCL_manuscript}

\end{document}